\documentclass[letterpaper]{article}

\usepackage{natbib,alifeconf}  
\usepackage{url,hyperref,cleveref}
\usepackage{booktabs}
\usepackage{multirow}
\usepackage{graphicx}
\usepackage{xcolor}

\newcommand\blfootnote[1]{%
  \begingroup
  \renewcommand\thefootnote{}\footnote{#1}%
  \addtocounter{footnote}{-1}%
  \endgroup
}

\title{CheeseBench: Evaluating Large Language Models\\on Rodent Behavioral Neuroscience Paradigms}

\author{
    Zacharie Bugaud$^{1}$ \\
    \mbox{}\\
    $^1$Astera Institute \\
    zacharie@astera.org
}

\begin{document}

\maketitle

\begin{abstract}
We introduce \textsc{CheeseBench}, a benchmark that evaluates large language models (LLMs) on nine classical behavioral neuroscience paradigms (Morris water maze, Barnes maze, T-maze, radial arm maze, star maze, operant chamber, shuttle box, conditioned place preference, and delayed non-match to sample), spanning six cognitive dimensions.
Each task is grounded in peer-reviewed rodent protocols with approximate animal baselines.
The agent receives a unified system prompt with no task-specific instructions and must discover goals purely from ASCII text observations and reward signals, much like a rodent placed into an unfamiliar apparatus.
We evaluate thirteen LLMs spanning six open-weight models (3B--72B parameters) and seven frontier proprietary models (Claude Haiku~4.5, Sonnet~4.6, Opus~4.6--4.7, GPT-4.1, GPT-5.2, GPT-5.2-Codex) and compare against both a random baseline and a graph-based reinforcement learning agent.
The best open-weight model (Qwen2.5-VL-7B) reaches 68.0\% average success on ASCII input versus 38.0\% for random agents and 78.9\% for approximate rodent baselines, while the best frontier model (Claude Opus~4.7) reaches 86.7\%, the first configuration in our evaluation to match or exceed the rodent reference.
We find that (1)~scaling beyond 7B yields diminishing returns within the open-weight Qwen2.5-VL family, (2)~longer context history degrades performance, (3)~chain-of-thought prompting hurts rather than helps, (4)~a vision-language architecture provides an advantage at 7B but hurts at 32B, and (5)~the working-memory bottleneck (Radial Arm Maze) that is nearly unsolved by open-weight models is largely closed by frontier models.
Because the same model's performance ranges from 20\% to 57\% depending on interface parameters alone, these results characterize the agent-plus-interface system, not the model in isolation.
Under this unified zero-shot ASCII protocol, open-weight LLM agents remain well below approximate rodent reference values, while frontier proprietary models close the gap, suggesting that the deficit is closable with sufficient scale and post-training rather than being a fundamental architectural limit.
\end{abstract}

Submission type: \textbf{Full Paper}
\blfootnote{Code: \url{https://github.com/stef41/CheeseBench} (also \texttt{pip install cheesebench}). Live leaderboard: \url{https://huggingface.co/spaces/zachz/cheesebench-leaderboard}. \textcopyright~2026 Zacharie Bugaud. Published under a Creative Commons Attribution 4.0 International (CC BY 4.0) license.}

\section{Introduction}

Large language models (LLMs) now perform well on question answering, reasoning, and multimodal benchmarks \citep{liu2024visual,team2024gemini}.
But these benchmarks primarily test declarative knowledge: recognizing objects, answering factual questions, or following explicit instructions.
A different kind of intelligence is procedural learning, i.e., discovering goals through trial and error, building spatial representations, and adapting behavior based on reward feedback.
This is what behavioral neuroscience has measured in laboratory animals for over a century.

Rodent behavioral paradigms provide standardized, reproducible assays of cognitive function \citep{vorhees2006morris,shoji2012t}.
Tasks like the Morris water maze and radial arm maze probe specific neural circuits: hippocampal place cells for spatial learning \citep{okeefe1971hippocampus}, working memory \citep{olton1976spatial}, and instrumental conditioning \citep{skinner1938behavior}.
These paradigms make natural AI benchmarks because they are (i)~mechanistically well-understood, (ii)~come with quantitative animal baselines from decades of research, and (iii)~test capabilities that current LLM benchmarks largely ignore.

We present \textsc{CheeseBench}, a benchmark suite of nine environments spanning six cognitive dimensions (\Cref{tab:environments}).
Each environment implements a published rodent protocol as a gridworld rendered as ASCII text (\Cref{fig:environments}).
Three rendering modes are available (ASCII top-down, egocentric, and pseudo-3D first-person), with ASCII top-down used as the default for all experiments.
The agent receives a single unified system prompt identical across all tasks, with no task-specific instructions, hints, or reward shaping.
It must discover the goal, form a strategy, and improve over repeated trials, just as a rodent placed in an unfamiliar apparatus would.
The primary goal of CheeseBench is to evaluate how well current LLM agent configurations perform on procedural discovery under a unified blind protocol; the rodent data serve as approximate biological reference points rather than strict competitive baselines.

Our contributions are:
\begin{enumerate}
    \item A benchmark suite of 9 behavioral neuroscience paradigms with approximate rodent reference values, covering 6 cognitive dimensions, released as an installable package (\texttt{pip install cheesebench}) with a public live leaderboard.
    \item A systematic evaluation of 13~LLMs --- six open-weight (3B--72B) and seven frontier proprietary models --- under a unified zero-shot ASCII protocol, showing that open-weight models plateau well below rodent references while frontier models match or exceed them.
    \item Ablation studies on the open-weight reference model showing that longer context hurts, chain-of-thought hurts, and the advantage of visual input over ASCII text is scale-dependent.
\end{enumerate}

\section{Related Work}

\paragraph{LLM Benchmarks.}
Standard LLM and VLM benchmarks such as VQA \citep{antol2015vqa}, MMLU \citep{hendrycks2021measuring}, and MathVista \citep{lu2024mathvista} primarily assess declarative reasoning.
Interactive benchmarks like ALFWorld \citep{shridhar2021alfworld} and WebArena \citep{zhou2024webarena} test procedural capabilities but use task-specific instructions and structured feedback.
\textsc{CheeseBench} uses a unified blind protocol with only scalar reward feedback.

\paragraph{Animal Cognition as AI Benchmark.}
The Animal-AI Testbed \citep{crosby2020animal} pioneered evaluating AI on animal cognition tests but focused on 3D physics (Unity engine) without direct rodent protocol fidelity.
Concurrent work has proposed bird cognition \citep{voudouris2023direct} and primate tests for LLMs.
Our work differs in that each environment maps to a specific published rodent protocol with approximate baselines derived from the peer-reviewed literature.

\paragraph{Cognitive Profiling of LLMs.}
Recent work has adapted human psychological batteries (Theory of Mind, analogical reasoning) to test LLMs \citep{kosinski2023theory,webb2023emergent}.
We complement this by using the nonverbal, reward-driven paradigms standard in comparative psychology.

\section{CheeseBench Benchmark}

\subsection{Design Principles}

\textsc{CheeseBench} follows four principles inspired by how behavioral neuroscience studies are conducted:

\begin{enumerate}
\item \textbf{Unified protocol.} A single system prompt for all nine tasks. No task-specific instructions, reward explanations, or goal descriptions. The agent must discover task structure through exploration and reward feedback.
\item \textbf{Approximate biological references.} Each environment corresponds to a real rodent experiment. We derive reference success rates from published learning curves by mapping the original outcome metric to our binary criterion (see the \emph{\nameref{app:baselines}} appendix). These are constructed reference anchors, not direct competitive baselines.
\item \textbf{Cognitive taxonomy.} Environments span six cognitive dimensions, enabling profiling across distinct neural substrates.
\item \textbf{Multi-action episodes.} The agent outputs up to $k$ actions per API call along with free-text ``learnings'' that serve as an explicit working memory, mimicking the episodic structure of animal experiments.
\end{enumerate}

\subsection{Environments}

\Cref{tab:environments} lists all nine paradigms.  We briefly describe representative tasks:

\paragraph{Morris Water Maze (MWM).}
A circular arena with a hidden escape platform. The agent uses distal landmarks for allocentric navigation. Rodents typically reach asymptotic performance within 5 sessions \citep{vorhees2006morris}.

\paragraph{Radial Arm Maze (RAM).}
An 8-arm maze with 4 baited arms. The agent must visit all baited arms without revisiting. This requires working memory across choices within each trial, and rodents learn to avoid revisits within several sessions \citep{olton1976spatial}.

\paragraph{Operant Chamber.}
A Skinner box with two levers and a magazine. The agent learns to press the correct lever to obtain reward via instrumental conditioning. Rodents rapidly acquire lever-press responses under continuous reinforcement \citep{skinner1938behavior}.

\paragraph{Shuttle Box.}
A two-compartment chamber with a conditioned stimulus (CS) preceding an aversive unconditioned stimulus (US). The agent must learn to shuttle to the other compartment during the CS to avoid the US. Rodents typically reach $\sim$70\% avoidance by session~5 \citep{chacon2016shuttle}.

\paragraph{Delayed Non-Match to Sample (DNMS).}
A touchscreen-style working-memory task in which the agent sees a sample stimulus, waits through a brief delay, and must then select the \emph{non-matching} option from a choice pair. We adapt the trial-unique non-matching-to-location (TUNL) variant used in rodent touchscreen studies; the rodent reference reflects 80\% correct on large spatial separations at short delay \citep{horner2013touchscreen}.

\begin{figure}[t]
\centering
\includegraphics[width=\columnwidth]{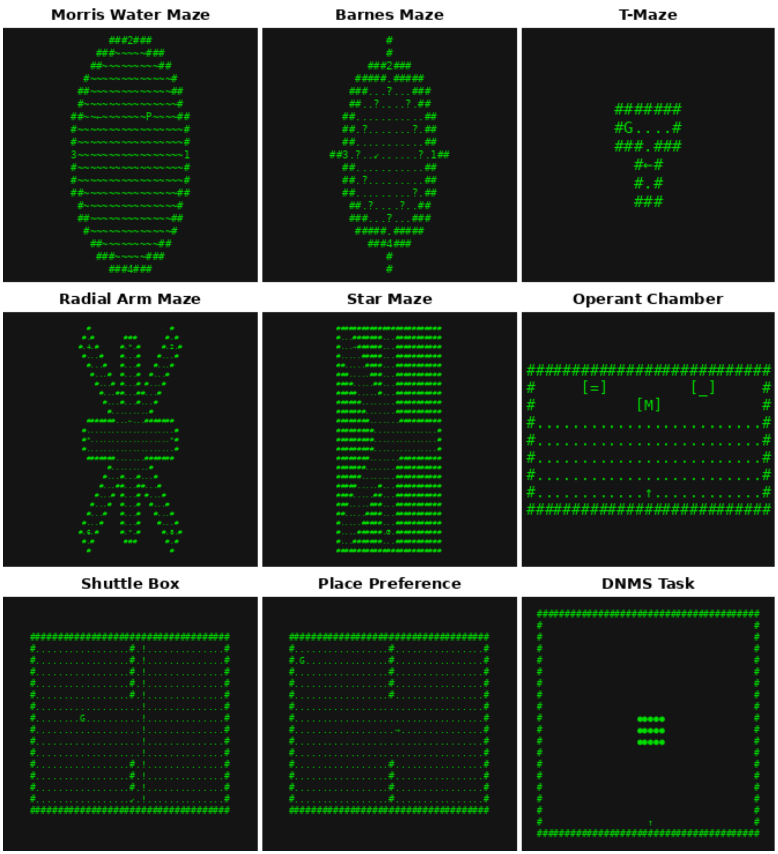}
\caption{ASCII renderings of all nine CheeseBench environments as seen by the model. Walls are shown as \texttt{\#}, traversable areas as \texttt{\~{}} or \texttt{.}, goals as \texttt{G}, and the agent arrow indicates position and facing direction.}
\label{fig:environments}
\end{figure}

\Cref{fig:environments} shows the ASCII rendering of all nine environments as the model sees them.

\begin{table}[t]
\centering
\caption{CheeseBench environments. Animal baselines are from published rodent studies. Trials = evaluation trials; Steps = max steps/trial.}
\label{tab:environments}
\resizebox{\columnwidth}{!}{%
\begin{tabular}{@{}llcrr@{}}
\toprule
Environment & Cognitive Dimension & Species & Trials & Steps \\
\midrule
MorrisWaterMaze & Spatial Learning & C57BL/6 & 20 & 500 \\
BarnesMaze & Spatial Learning & C57BL/6 & 16 & 300 \\
StarMaze & Spatial Learning & C57BL/6 & 40 & 300 \\
TMaze & Ego. Navigation & C57BL/6 & 40 & 200 \\
RadialArmMaze & Working Memory & C57BL/6 & 20 & 400 \\
DNMSTask & Working Memory & Long-Evans & 100 & 50 \\
OperantChamber & Instr. Conditioning & C57BL/6 & 50 & 100 \\
ShuttleBox & Avoidance Learning & SD rat & 40 & 50 \\
PlacePreference & Assoc. Learning & C57BL/6 & 12 & 300 \\
\bottomrule
\end{tabular}%
}
\end{table}

\subsection{Observation and Action Space}

\begin{figure}[t]
\centering
\includegraphics[width=\columnwidth]{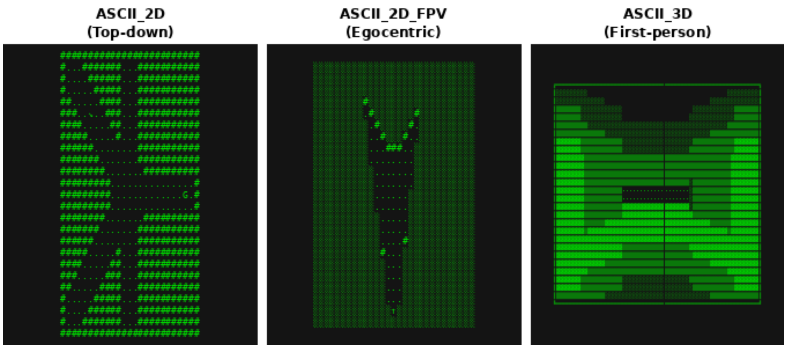}
\caption{The three ASCII observation modes for the Star Maze. \texttt{ASCII\_2D} provides a full top-down text map; \texttt{ASCII\_2D\_FPV} shows an egocentric cropped view centered on the agent; \texttt{ASCII\_3D} renders a pseudo-3D first-person perspective with depth shading.}
\label{fig:viewmodes}
\end{figure}

Environments are rendered as ASCII text in three view modes (\Cref{fig:viewmodes}): \texttt{ASCII\_2D} (top-down bird's-eye map showing the full layout), \texttt{ASCII\_2D\_FPV} (egocentric cropped view centered on the agent), and \texttt{ASCII\_3D} (pseudo-3D first-person perspective using block shading).
The default mode used in all experiments is \texttt{ASCII\_2D}.
The action space consists of four egocentric actions: \texttt{FORWARD}, \texttt{ROTATE\_LEFT}, \texttt{ROTATE\_RIGHT}, and \texttt{STAY}.
At each turn, the agent produces up to $k$ actions and a free-text ``learnings'' field for explicit working memory, capped at 500 characters.

\subsection{System Prompt}

The unified system prompt describes the action space and output format but provides \emph{no} information about task goals, reward structure, or environment layout.
It also includes generic strategy guidance (attend to reward signals, track spatial position, form and test hypotheses, plan efficiently), which is identical across all tasks:

\begin{quote}
\itshape
``You are an embodied agent placed in a behavioral experiment. Your only goal is to maximize cumulative reward. [\ldots] STRATEGY: 1.~Reward signals: positive $\to$ repeat, negative $\to$ change. 2.~Spatial memory: track position. 3.~Hypothesis testing. 4.~Pattern recognition. 5.~Efficient planning. [\ldots] Respond with LEARNINGS and up to $k$ ACTIONS.''
\end{quote}

\section{Experimental Setup}

\paragraph{Models.}
We evaluate two complementary tiers of LLMs.
\textbf{Open-weight tier:} six models from three families served locally via vLLM on 8$\times$H100 80GB GPUs --- \textbf{Qwen2.5-VL} at 3B, 7B, 32B, and 72B parameters \citep{bai2025qwen25}, \textbf{InternVL2.5-8B} \citep{chen2024internvl}, and \textbf{Phi-4-Multimodal-14B} (abbreviated Phi-4-MM-14B) \citep{abdin2024phi4}. Although these are vision-language models by architecture, the main experiments use \texttt{ASCII\_2D} text rendering, so the models receive only text input.
\textbf{Frontier tier:} seven proprietary models accessed through a uniform OpenAI-compatible interface --- \textbf{Claude Haiku~4.5, Sonnet~4.6, Opus~4.6, Opus~4.7} \citep{anthropic2026claude}, and \textbf{GPT-4.1, GPT-5.2, GPT-5.2-Codex} \citep{openai2026gpt52}. The frontier evaluations use the identical unified system prompt and the same three view modes as the open-weight tier, but a smaller per-environment trial count (10 trials/env $\times$ 9 envs $\times$ 3 view modes = 270 trials per model; GPT-4.1 is run at 3 trials/env due to its pathologically high call rate).
For the ablation analyses we use the open-weight reference model (Qwen2.5-VL-7B) for cost reasons.

\paragraph{Protocol.}
Each model runs all 9 environments across the three ASCII view modes (\texttt{ASCII\_2D}, \texttt{ASCII\_2D\_FPV}, \texttt{ASCII\_3D}) with default interface settings: history length $h{=}5$ observation-action pairs, action batch size $k{=}8$, default prompt variant.
The environment state is rendered as an ASCII character grid and sent as a text message; no images are involved.
Unless otherwise noted, the per-model ``overall'' success rate we report is the mean across environments of the best per-environment view mode (i.e., for each environment we take the best of the three ASCII view modes for that model, then average over environments).
This matches the metric used on the public CheeseBench leaderboard. ASCII\_2D-only numbers are tabulated in the \emph{\nameref{app:ascii2d}} appendix.
We compare against two non-LLM baselines: a uniformly random agent and Tabular-QL, a reinforcement learning agent that hashes observations into discrete states, learns via tabular Q-learning, and builds a transition graph for BFS planning to previously discovered goals.
We also test a BFS oracle that parses the ASCII grid and navigates to any visible goal marker; this agent reaches 100\% success on tasks where the goal is visible in the grid (e.g., Morris Water Maze, where the hidden platform is rendered as a symbol visible to any text parser but requires spatial learning for the LLM agent to exploit).
We report success rates across all trials per environment; error bars in \Cref{fig:main_results} show Wilson 95\% binomial confidence intervals computed from the per-environment trial counts in \Cref{tab:environments} (12--100 trials depending on the task).
All runs use temperature~0.7 sampling with fixed environment seeds; stochastic environment dynamics and sampling are the sources of run-to-run randomness.
Each model--environment condition is a single run; we do not average over multiple seeds, so per-environment estimates carry non-trivial sampling uncertainty.
The approximate rodent values shown as diamond markers in figures are not measured in our environments; they are reference anchors derived from published learning curves under different protocols and outcome metrics (see the \emph{\nameref{app:baselines}} appendix).

\paragraph{Ablations.}
Using Qwen2.5-VL-7B as the reference model, we ablate:
\begin{itemize}
    \item \textbf{History length} $h \in \{1, 3, 5, 10\}$: number of past observation-action pairs retained in context.
    \item \textbf{Prompt variant}: default, minimal (stripped-down), chain-of-thought (CoT), and few-shot (with example trajectories).
    \item \textbf{Action batch size} $k \in \{1, 4, 8, 16\}$: number of actions per API call.
    \item \textbf{Vision}: VLM with image input vs.\ text-only LLM (Qwen2.5-Instruct without vision encoder) at 7B and 32B scales, both using ASCII text observations. This tests whether the vision encoder helps even when the input is text.
\end{itemize}

\section{Results}

\subsection{Open-Weight Multi-Model Comparison}

\begin{figure}[t]
\centering
\includegraphics[width=\columnwidth]{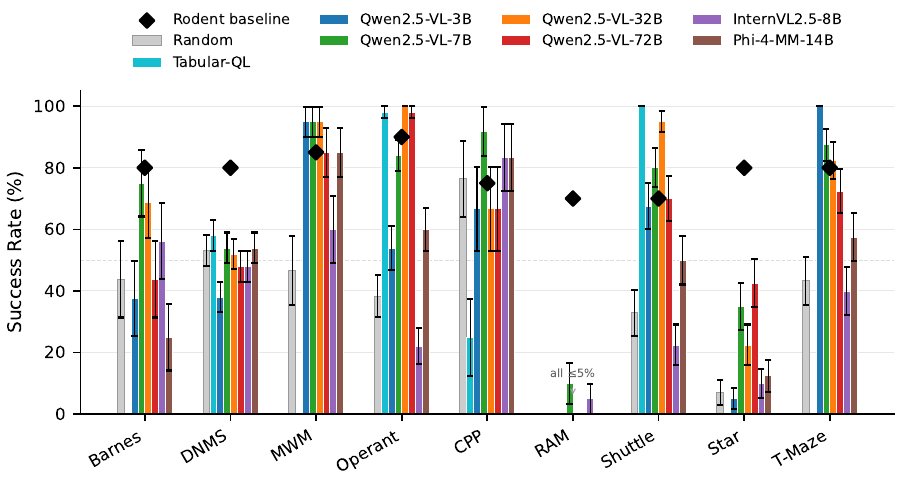}
\caption{Per-environment success rates (ASCII text input, best of three view modes per environment) for six open-weight LLMs, Tabular-QL, and random baseline compared to approximate rodent reference values (diamonds). The Radial Arm Maze remains essentially unsolved by all open-weight agents; the \emph{Frontier Models} subsection shows that frontier models close this gap.}
\label{fig:main_results}
\end{figure}

\Cref{fig:main_results} shows per-environment success rates for the six open-weight models on ASCII text input (best of three view modes per environment).
The best overall open-weight model is Qwen2.5-VL-7B (68.0\%), followed by Qwen2.5-VL-32B (64.7\%) and Qwen2.5-VL-72B (58.5\%).
The random baseline averages 38.0\% and the Tabular-QL agent averages 31.2\%, but they succeed on very different tasks: Tabular-QL reaches 98\% on OperantChamber and 100\% on ShuttleBox through Q-learning, while scoring 0\% on all maze-based spatial environments.
All open-weight LLMs fall well below the approximate rodent reference values (70--90\%), shown as diamond markers; we revisit this gap for frontier models in the \nameref{sec:frontier} subsection below.

Performance varies widely across environments.
On \textbf{OperantChamber} (instrumental conditioning), Qwen2.5-VL-32B reaches 100\%, matching the approximate rodent reference value, which suggests that simple stimulus-response associations are within reach of in-context learning.
On \textbf{ShuttleBox}, Qwen2.5-VL-32B reaches 95\% --- the strongest open-weight result on this task.
But \textbf{RadialArmMaze} (working memory) is nearly unsolved by open-weight models: the best (Qwen2.5-VL-7B) manages 10.0\%, compared to 70\% for rodents.
This task requires keeping track of which arms have been visited across 20+ steps within a single trial, which is very difficult for autoregressive models at this scale.

\subsection{Cognitive Profile}

\begin{figure}[t]
\centering
\includegraphics[width=0.75\columnwidth]{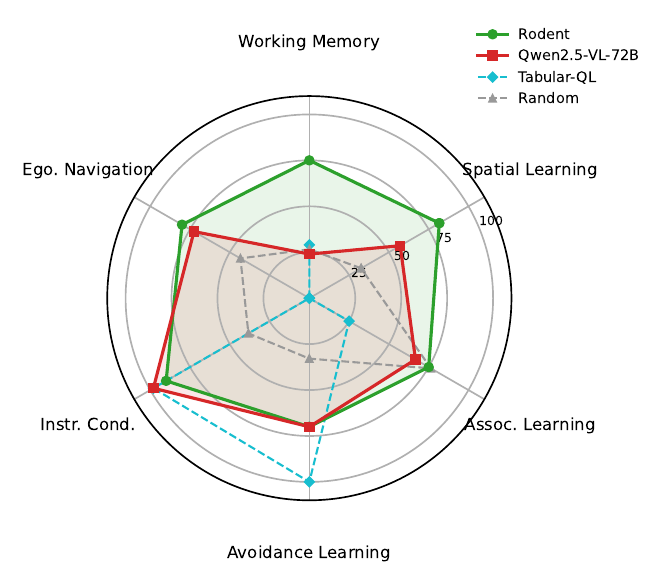}
\caption{Cognitive profile across six dimensions. The LLM and Tabular-QL agent show complementary strengths: the LLM handles a broader range of tasks while Tabular-QL excels at conditioning through direct reinforcement learning.}
\label{fig:radar}
\end{figure}

\Cref{fig:radar} shows the cognitive profile of the largest open-weight Qwen model (Qwen2.5-VL-72B) compared to rodent reference values, as it displays the clearest contrast across cognitive dimensions among open-weight models.
It reaches 98\% on instrumental conditioning, where simple reward associations suffice, but collapses to 0\% on the working-memory task RAM (vs.\ 70\% for rodents).
Spatial learning (averaging across MWM, Barnes and Star: $\sim$57\%) and avoidance learning (Shuttle: 70\%) fall in between.
The pattern suggests that open-weight LLMs in the 3B--72B range can learn reactive associations through in-context learning but struggle with tasks requiring sustained state tracking; the \nameref{sec:frontier} subsection shows that frontier models substantially close the working-memory gap.

\subsection{Scaling Behavior (Open-Weight)}

\begin{figure}[t]
\centering
\includegraphics[width=0.75\columnwidth]{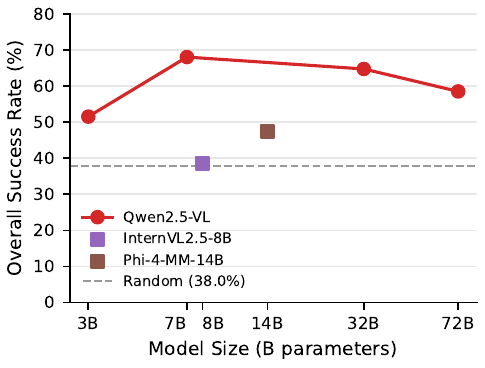}
\caption{Scaling behavior: performance saturates beyond 7B under this protocol.}
\label{fig:scaling}
\end{figure}

\Cref{fig:scaling} shows overall success rate vs.\ model size within the Qwen2.5-VL family.
There is a large jump from 3B (51.5\%) to 7B (68.0\%), but performance plateaus beyond 7B, with 32B (64.7\%) and 72B (58.5\%) performing similarly or slightly worse; differences between the 7B, 32B, and 72B models are within Wilson 95\% confidence intervals.
This saturation contrasts with typical benchmark scaling laws.
InternVL2.5-8B (38.6\%) and Phi-4-MM-14B (47.5\%) underperform the comparable-sized Qwen models, so model family and training data clearly matter too.
Frontier proprietary models break out of this plateau (see \nameref{sec:frontier} below), suggesting the saturation reflects post-training and RL-with-tools recipes rather than parameter count alone.

\subsection{Ablation Studies}

\begin{figure}[t]
\centering
\includegraphics[width=\columnwidth]{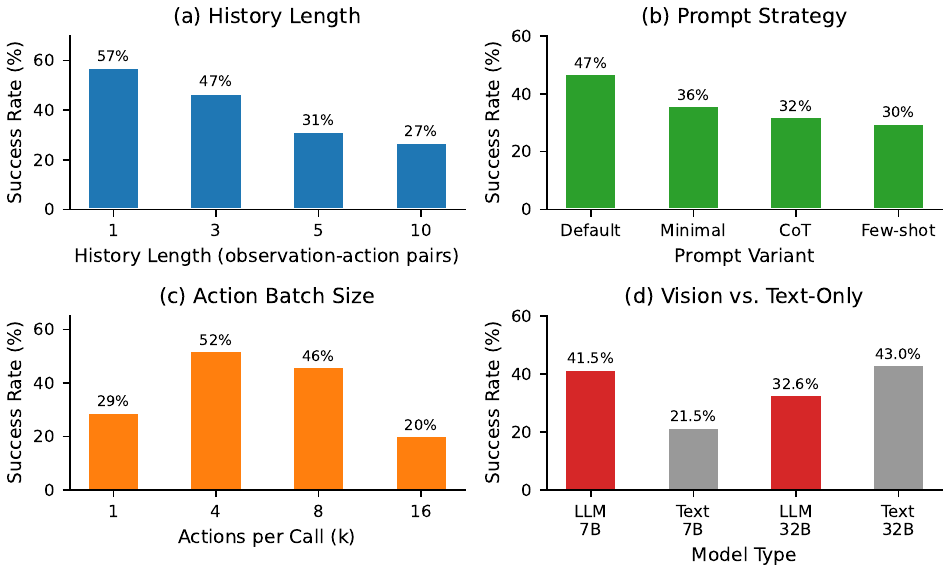}
\caption{Ablation studies (all on ASCII text input) using Qwen2.5-VL-7B (32B for vision ablation). (a)~Shorter history is better. (b)~Default prompt beats CoT. (c)~$k{=}4$ is the sweet spot. (d)~Vision encoder helps at 7B but hurts at 32B.}
\label{fig:ablations}
\end{figure}

\Cref{fig:ablations} summarizes the four ablation dimensions.

\emph{Note on ablation protocol:} to manage compute, ablation runs use a flat 15 trials per environment rather than the per-environment protocol in \Cref{tab:environments}, and each condition is a separate run.
With only 15 trials, per-condition estimates carry substantial sampling noise (binomial 95\% CI of $\pm$25pp at $p{=}0.5$), so the ``default'' ablation baseline ($h{=}5$, $k{=}8$, default prompt) need not exactly match the 68.0\% multi-model result for the same model.
Trends should be interpreted directionally; individual effect sizes are approximate.

\paragraph{History length (\Cref{fig:ablations}a).}
Shorter history tends to give better results: $h{=}1$ achieves 57.0\% vs.\ 26.7\% for $h{=}10$, though with 15 trials per environment the absolute values are noisy (see protocol note above).
The directional trend is consistent with the model having trouble picking out relevant past observations from longer context.
With $h{=}1$, the ``learnings'' field (500 characters of free-text working memory) becomes the only long-term memory, and the model appears to use it more effectively.

\paragraph{Prompt strategy (\Cref{fig:ablations}b).}
The default prompt (46.7\%) outperforms chain-of-thought (31.9\%), minimal (35.6\%), and few-shot (29.6\%).
CoT hurting is worth noting: step-by-step reasoning appears to waste context tokens on verbose output that displaces useful observation history.
Few-shot examples may bias the model toward mimicking the demonstrated trajectories instead of adapting to new environments.

\paragraph{Action batch size (\Cref{fig:ablations}c).}
Performance is highest at $k{=}4$ (51.9\%), followed by $k{=}8$ (45.9\%), with poor results at the extremes, $k{=}1$ (28.9\%) and $k{=}16$ (20.0\%).
With $k{=}1$, the API call overhead means fewer observations per episode.
With $k{=}16$, the model must plan 16 steps ahead without feedback, which is too long for reliable open-loop control.

\paragraph{Vision vs.\ text-only (\Cref{fig:ablations}d).}
This ablation compares VLM architectures (with vision encoder) against text-only LLMs (without vision encoder), both receiving ASCII text.
At 7B, the VLM (41.5\%) outperforms text-only (21.5\%) by 20 percentage points.
While this gap is consistent with the vision encoder providing an advantage even on text input, we cannot rule out alternative explanations such as differences in training data or fine-tuning procedures between the VLM and text-only variants.
At 32B, the pattern reverses: the VLM (32.6\%) underperforms text-only (43.0\%) by 10.4 percentage points, suggesting that at larger scales the vision encoder may interfere with text-based reasoning rather than help.

\subsection{Frontier Models}
\label{sec:frontier}

We additionally evaluate seven frontier proprietary models accessed through a uniform OpenAI-compatible interface (Claude Haiku~4.5, Sonnet~4.6, Opus~4.6--4.7, GPT-4.1, GPT-5.2, GPT-5.2-Codex), under the same unified system prompt and same three view modes.
\Cref{tab:frontier} summarizes per-environment success rates.

Three findings stand out:
\textbf{(1)~Frontier models close the rodent gap.}
Claude Opus~4.7 reaches 86.7\% overall, the first configuration in our evaluation to match or exceed the 78.9\% rodent reference average; GPT-5.2-Codex (82.2\%), Claude Opus~4.6 (82.2\%), and Claude Sonnet~4.6 (81.1\%) also exceed the rodent average. The best open-weight model trails by 13--19 percentage points.
\textbf{(2)~Working memory becomes solvable.}
The Radial Arm Maze, nearly unsolved by open-weight models (best 10\%), is solved at 100\% by Claude Opus~4.7 and 90\% by Claude Opus~4.6 --- evidence that the within-trial state-tracking bottleneck is not architectural to transformers but is closable by sufficiently capable post-trained models.
GPT-4.1 is a striking counter-example: it achieves 0\% on RAM and exhibits a strong action bias (predominantly emitting \texttt{STAY}) that inflates its call count by roughly 3$\times$ relative to peers.
\textbf{(3)~Avoidance learning regresses.}
The ShuttleBox task, on which open-weight Qwen2.5-VL-32B reaches 95\%, drops to 10--40\% for most frontier models (only Claude Haiku~4.5 reaches 60\%). The pattern is consistent with frontier models being more cautious about emitting actions early under noisy reward, which delays the conditioning curve relative to the 50-step trial budget. This is a candidate failure mode worth further study.

These frontier results were collected at 10 trials per environment per view mode (270 trials per model; 81 trials for GPT-4.1 due to its inflated call cost), which is fewer than the open-weight runs (typically 50 trials per environment per view mode). Per-cell Wilson 95\% CIs are correspondingly wider; we report the full per-environment table in the appendix and on the live leaderboard.

\begin{table}[t]
\centering
\caption{Frontier-model overall success rate (\%) compared to the best open-weight model and the rodent reference. ``Overall'' is the mean of per-environment best-view-mode success rate. Trial counts: frontier models 270 per run (GPT-4.1: 81); open-weight 1352 per run.}
\label{tab:frontier}
\footnotesize
\begin{tabular}{@{}lr@{}}
\toprule
\textbf{Model (tier)} & \textbf{Overall} \\
\midrule
Claude Opus~4.7 (frontier) & \textbf{86.7} \\
GPT-5.2-Codex (frontier) & 82.2 \\
Claude Opus~4.6 (frontier) & 82.2 \\
Claude Sonnet~4.6 (frontier) & 81.1 \\
GPT-5.2 (frontier) & 75.6 \\
Claude Haiku~4.5 (frontier) & 70.0 \\
Qwen2.5-VL-7B (open-weight, best) & 68.0 \\
GPT-4.1 (frontier) & 55.6 \\
\midrule
Rodent reference (approx.) & 78.9 \\
\bottomrule
\end{tabular}
\end{table}

\subsection{Image-Based Rendering}

We also evaluated the pixel-based \texttt{TOPDOWN\_2D} rendering on the same six open-weight VLMs.
Qwen2.5-VL-32B (vision) reached 50.2\% overall and Qwen2.5-VL-7B (vision) reached 43.0\%, both below their text-input counterparts (64.7\% and 68.0\% respectively). Smaller open-weight VLMs (Qwen2.5-VL-3B, Phi-4-MM-14B, InternVL2.5-8B) score 24--31\% in vision mode, well below their text scores.
We attribute the gap to image-renderer artefacts (the current renderer produces noisy thumbnails) rather than a fundamental modality limit; full per-environment vision-mode numbers are on the live leaderboard.

\begin{table}[t]
\centering
\caption{Numerical summary. Overall success rate (\%) on ASCII text input, best of three view modes per environment. Multi-model rows use per-environment trial counts (\Cref{tab:environments}); ablations use 15~trials/env in separate runs and may therefore differ from multi-model values. Frontier models are reported in \Cref{tab:frontier}.}
\label{tab:summary}
\begin{minipage}[t]{0.48\columnwidth}
\centering
\footnotesize
\textbf{Open-weight \& baselines}\\[2pt]
\begin{tabular}{@{}lr@{}}
\toprule
Model & \% \\
\midrule
Qwen2.5-VL-7B  & 68.0 \\
Qwen2.5-VL-32B & 64.7 \\
Qwen2.5-VL-72B & 58.5 \\
Qwen2.5-VL-3B  & 51.5 \\
Phi-4-MM-14B   & 47.5 \\
InternVL2.5-8B & 38.6 \\
Random         & 38.0 \\
Tabular-QL     & 31.2 \\
\bottomrule
\end{tabular}
\end{minipage}\hfill
\begin{minipage}[t]{0.48\columnwidth}
\centering
\footnotesize
\textbf{Ablations (Qwen2.5-VL-7B)}\\[2pt]
\begin{tabular}{@{}lr@{}}
\toprule
Condition & \% \\
\midrule
History $h{=}1$    & 57.0 \\
History $h{=}3$    & 46.7 \\
History $h{=}5$    & 31.1 \\
History $h{=}10$   & 26.7 \\
Actions $k{=}4$    & 51.9 \\
Actions $k{=}8$    & 45.9 \\
Prompt: default   & 46.7 \\
Prompt: CoT       & 31.9 \\
Vision LLM-7B     & 41.5 \\
Vision Text-7B    & 21.5 \\
Vision LLM-32B    & 32.6 \\
Vision Text-32B   & 43.0 \\
\bottomrule
\end{tabular}
\end{minipage}
\end{table}

\section{Discussion}

The results reveal a coherent pattern across all experiments.
We first summarize the empirical findings, then discuss possible interpretations.

\paragraph{Findings.}
(1)~Radial Arm Maze is nearly unsolved by all open-weight agents (0--10\%), while frontier models such as Claude Opus~4.7 (100\%) and Claude Opus~4.6 (90\%) crack it; conditioning tasks (OperantChamber) approach or match rodent reference values across most tiers.
(2)~Within the open-weight Qwen2.5-VL family, performance jumps from 3B to 7B but does not improve further at 32B or 72B; the open-weight plateau is broken by frontier proprietary models.
(3)~Shorter history ($h{=}1$) outperforms longer history ($h{=}10$) by roughly 30 percentage points in the ablation, though estimates are noisy (15 trials/env).
(4)~Chain-of-thought prompting hurts.
(5)~$k{=}4$ actions per call is the sweet spot; extreme values degrade performance.
(6)~The avoidance-learning task (ShuttleBox) shows a surprising regression in the frontier tier (10--40\% for most models, 60\% for Claude Haiku~4.5) relative to the open-weight Qwen-VL models ($\sim$95\%), a candidate failure mode worth dedicated study.
These patterns are robust across models within each tier.

\paragraph{Interpretations.}
The findings are consistent with several possible explanations, which we present as hypotheses rather than conclusions.

\emph{State-tracking bottleneck.}
The near-total failure of open-weight LLMs on RAM and the inverted history-length result both point toward difficulty maintaining and updating internal state representations at the 3B--72B scale.
In biological agents, hippocampal place cells and prefrontal working memory circuits support within-trial state tracking automatically.
In our open-weight LLM agents, the only state-tracking mechanism is the 500-character ``learnings'' scratchpad, updated only between API calls.
That frontier models (Claude Opus~4.7, 4.6) solve RAM at 90--100\% under the \emph{same} interface and scratchpad indicates the bottleneck is not architectural to transformers, nor inherent to this prompting interface, but reflects capability differences in post-training, scale, and likely RL-with-tools recipes accessible to frontier-lab models.

\emph{Context management.}
The inverted history-length result ($h{=}1 > h{=}10$) suggests that the model cannot selectively attend to relevant past observations when mixed with stale ones.
With $h{=}1$, the learnings field becomes the sole persistent memory, and the model appears to use it more effectively as a compressed summary.
Again, this may reflect the specific prompting interface as much as an inherent limitation.

\emph{Scaling saturation.}
The plateau beyond 7B \emph{within the open-weight Qwen2.5-VL family} contrasts with typical NLP benchmark scaling.
One interpretation is that for procedural behavioral tasks, the bottleneck is not world knowledge or language understanding but the ability to plan under uncertainty given the information available in the prompt.
The frontier results suggest that scale alone is not the limiting factor either: capability gains from frontier-lab post-training pipelines, not parameter count, appear to drive the additional 13--19 percentage points over the best open-weight model.

\paragraph{Agent-Plus-Interface Evaluation.}
Since observed performance is tightly coupled to interface design (history length, action batching, scratchpad size, prompt variant), CheeseBench should be understood as evaluating the \emph{agent-plus-interface system}, not raw latent model cognition.
The ablation results demonstrate this clearly: the same model's performance ranges from 20\% to 57\% depending on interface parameters alone.
This framing does not weaken the benchmark; it sharpens the question from ``how smart is the model?'' to ``how well does this agent configuration perform on procedural discovery tasks?''

\paragraph{Limitations.}
The comparison between LLM and rodent performance is not fully symmetric: animal baselines are approximate values derived from published learning curves and represent asymptotic performance after multiple training sessions, while our LLM agents are evaluated zero-shot within a single benchmark run under a specific prompting protocol and interface.
The gap we report is therefore specific to this protocol; a different interface (e.g., longer scratchpad, per-step memory updates, or a fine-tuned policy) could narrow it.
The environments are simplified gridworlds that lack the sensory richness of real rodent arenas; we discuss what is preserved and what is lost in the \emph{\nameref{app:fidelity}} appendix.
Additionally, in the Morris Water Maze implementation, the hidden platform is rendered as a visible symbol in the ASCII grid; the BFS oracle exploits this, and the 100\% BFS score on MWM reflects symbol-parsing ability rather than spatial search.
LLM agents also occasionally produce unparseable outputs (e.g., invalid action strings), which are counted as wasted steps; the rate varies across models (from near-zero for larger models to $\sim$0.5 per trial for 3B models) and represents a confound separate from cognitive ability.
We evaluate open-weight models only; proprietary models may perform differently.
We also do not test interventions (e.g., external memory, recurrence, or fine-tuning) that could directly probe hypothesized bottlenecks; such experiments are a natural next step.

\section{Conclusion}

Under a unified zero-shot ASCII protocol, open-weight LLM agents (3B--72B) remain well below approximate rodent reference values on standardized behavioral paradigms, while frontier proprietary models (Claude Opus~4.x, GPT-5.2-Codex) match or exceed those references.
The benchmark exposes a structured pattern within the open-weight tier --- LLM agents handle simple conditioning but struggle with spatial navigation and within-trial state tracking --- and a new pattern in the frontier tier, where the working-memory gap (RAM) is largely closed but a surprising regression appears on avoidance learning (ShuttleBox).
Because performance is tightly coupled to interface design, these results characterize the agent-plus-interface system rather than raw model cognition.
That the open-weight gap on RAM closes entirely under the same interface for frontier models suggests the bottleneck is not architectural and that closing remaining gaps will involve a combination of scale, post-training recipes, and interface design rather than any single intervention.
We hope this work encourages the Artificial Life community to explore memory-augmented agent architectures and to use behavioral neuroscience as a grounding discipline for evaluating artificial cognition.

The benchmark is publicly released as the \texttt{cheesebench} package on PyPI (\texttt{pip install cheesebench}); source at \url{https://github.com/stef41/CheeseBench}. A live leaderboard tracking new model releases is hosted at \url{https://huggingface.co/spaces/zachz/cheesebench-leaderboard}, with per-model report cards under \texttt{cards/} in the repository.

\bibliographystyle{apalike}
\bibliography{references}

\appendix
\section{ASCII\_2D-Only Results}
\label{app:ascii2d}

The headline numbers in the main text use the mean of per-environment best-view-mode success rates (matching the public leaderboard). For completeness, raw per-trial data for all (model, environment, view mode) cells is published with the source distribution under \texttt{results/} and on the live leaderboard. Restricting to \texttt{ASCII\_2D} only (the original paper protocol) lowers headline numbers by approximately 5--15 percentage points per model without changing the overall ranking among open-weight models or the relative ordering between open-weight and frontier tiers.

\section{Environment Fidelity}
\label{app:fidelity}

\Cref{tab:fidelity} summarizes what each CheeseBench environment preserves from the original rodent protocol and what is simplified.
The core simplification shared by all tasks is the reduction to a discrete gridworld with four egocentric actions and ASCII text rendering.
Despite these abstractions, each environment retains the essential cognitive demand of the source paradigm: spatial navigation tasks still require allocentric or egocentric mapping, working memory tasks still require tracking visited locations across steps, and conditioning tasks still require associating stimuli with rewards or punishments from scalar feedback alone.

\begin{table*}[t]
\centering
\caption{Environment fidelity summary. \textbf{P}: preserved from rodent protocol. \textbf{S}: simplified away. \textbf{C}: cognitive demand retained.}
\label{tab:fidelity}
\footnotesize
\begin{tabular}{@{}lp{14cm}@{}}
\toprule
\textbf{Environment} & \textbf{P / S / C} \\
\midrule
MWM & \textbf{P:} Hidden goal, distal landmarks, multiple starts. \textbf{S:} Continuous swim $\to$ grid; no currents. \textbf{C:} Allocentric spatial learning. \\[3pt]
Barnes & \textbf{P:} Escape hole among decoys, aversive motivation. \textbf{S:} Light/noise $\to$ step penalty; 3D $\to$ 2D. \textbf{C:} Spatial reference memory. \\[3pt]
Star & \textbf{P:} Radial corridors, strategy choice. \textbf{S:} Hidden goal $\to$ visible marker; continuous motion $\to$ grid; corridor width. \textbf{C:} Spatial strategy selection. \\[3pt]
T-Maze & \textbf{P:} T-corridor, forced choice, reward at one arm. \textbf{S:} Olfactory/tactile cues; continuous turning. \textbf{C:} Egocentric navigation. \\[3pt]
RAM & \textbf{P:} 8 arms (4 baited), revisit tracking. \textbf{S:} Odor trails; arm length variation; no explicit revisit penalty. \textbf{C:} Working memory across choices. \\[3pt]
DNMS & \textbf{P:} Sample--delay--choice, non-match rule. \textbf{S:} Object features $\to$ symbols; step-based delay. \textbf{C:} Working memory + rule learning. \\[3pt]
Operant & \textbf{P:} Levers, magazine, free-operant schedule. \textbf{S:} Pellet $\to$ scalar reward; no hunger drive. \textbf{C:} Instrumental conditioning. \\[3pt]
Shuttle & \textbf{P:} Two compartments, CS$\to$US, avoidance. \textbf{S:} Auditory CS $\to$ text; shock $\to$ neg.\ reward. \textbf{C:} Avoidance learning. \\[3pt]
CPP & \textbf{P:} Two chambers, one reward-paired, preference test. \textbf{S:} Drug/food $\to$ reward signal; floor $\to$ ASCII. \textbf{C:} Associative place learning. \\
\bottomrule
\end{tabular}
\end{table*}

\section{Rodent Baseline Construction}
\label{app:baselines}

\Cref{tab:baselines} documents how each approximate rodent baseline was derived.
Baselines are not single numbers extracted from one paper; they are approximate asymptotic success rates estimated from published learning curves for the relevant paradigm and species.
We mapped each paradigm's standard outcome metric (escape latency, error count, lever-press rate, etc.) to a binary success criterion appropriate for the gridworld implementation, then estimated the proportion of trials meeting that criterion at the end of training based on the shape of published learning curves.
These values should be treated as order-of-magnitude reference points, not precise quantitative benchmarks.

\begin{table*}[t]
\centering
\caption{Rodent baseline construction. Approx.\ = approximate asymptotic success rate used in figures. Sessions = number of training sessions modeled in the learning curve. PMC = PubMed Central source article for protocol.}
\label{tab:baselines}
\footnotesize
\begin{tabular}{@{}lcclp{10cm}@{}}
\toprule
\textbf{Env} & \textbf{Approx.} & \textbf{Sess.} & \textbf{PMC} & \textbf{Notes} \\
\midrule
MWM & 85\% & 5 & PMC3259155 & Platform acquisition, C57BL/6 mice, 5 trials/sess. Metric is path length; mapped to binary (reach platform within step limit). \\[2pt]
Barnes & 80\% & 5 & PMC1783636 & 12 holes, 4 trials/sess. Metric: primary errors $\to$ binary correct-hole. \\[2pt]
Star & 80\% & 10 & PMC3695082 & 5-arm maze, 4 trials/sess. Metric: reach goal arm via correct path. \\[2pt]
T-Maze & 80\% & 4 & PMC3399492 & Forced alternation, 10 trials/sess. Metric: correct arm on free choice. \\[2pt]
RAM & 70\% & 6 & PMC4030456 & 8 arms, 4 baited. Collect all rewards with 0 WM errors. Conservative; original rats scored higher on simpler metrics. \\[2pt]
DNMS & 80\% & 30 & PMC3982138 & Touchscreen TUNL task (rats). Criterion is 80\% correct on large separations at short delay. \\[2pt]
Operant & 90\% & 5 & PMC6619163 & FR-1 schedule, C57BL/6 mice, 20-min sess. Metric: lever press within step limit. Value reflects post-criterion steady-state performance; mean sessions to criterion is 6.5. \\[2pt]
Shuttle & 70\% & 5 & PMC4633642 & Active avoidance, Sprague-Dawley rats, 30 trials/sess. Metric: shuttle before US onset. Value from exercised group; sedentary controls reach $\sim$53\%. \\[2pt]
CPP & 75\% & 6 & PMC6101638 & 3-phase protocol. Metric: $>$55\% time in reward-paired chamber. \\
\bottomrule
\end{tabular}
\end{table*}

\end{document}